\documentclass{article}

\usepackage{PRIMEarxiv}

\usepackage{hyperref}
\usepackage{orcidlink}
\usepackage{url}
\usepackage{booktabs} % professional-quality tables
\usepackage[utf8]{inputenc} % allow utf-8 input
\usepackage[T1]{fontenc}    % use 8-bit T1 fonts
\usepackage{amsmath,amsfonts,amssymb}
\usepackage{microtype}
\usepackage{fancyhdr}
\usepackage{float}
\usepackage{cite}
\usepackage{graphicx} % For including images

\title{RACE-Align: Retrieval-Augmented and Chain-of-Thought Enhanced Preference Alignment for Large Language Models}

\author{
  Qihang Yan \orcidlink{0009-0002-0543-5620} \\
  ShanghaiTech University \\
  \texttt{yanqh2022@shanghaitech.edu.cn} \\
  \And
  Xinyu Zhang \orcidlink{0009-0003-1393-9500} \\
  Henan University \\
  \texttt{xinyu-zhang@henu.edu.cn} \\
  \And
  Luming Guo \orcidlink{0009-0003-3409-4578} \\
  Henan University \\
  \texttt{luming-guo@outlook.com} \\
  \And
  Qi Zhang \orcidlink{0009-0001-9305-9487} \\
  Liaoning University of Traditional Chinese Medicine \\
  \texttt{zhangqiliaoning@hotmail.com} \\
  \And
  Feifan Liu\thanks{Corresponding author} \orcidlink{0009-0009-7592-0999} \\
  Henan University \\
  \texttt{feifan-liu@henu.edu.cn} \\
}

\begin{document}
\maketitle

\begin{abstract}
    Large Language Models (LLMs) struggle with accuracy, domain-specific reasoning, and interpretability in vertical domains. Traditional preference alignment methods like Reinforcement Learning from Human Feedback (RLHF) and Direct Preference Optimization (DPO) often overlook the underlying knowledge sources and reasoning logic. This paper introduces RACE-Align (Retrieval-Augmented and Chain-of-Thought Enhanced Alignment), a novel framework designed to address these limitations. RACE-Align systematically constructs a binary preference dataset incorporating external knowledge support and explicit Chain-of-Thought (CoT) reasoning, then aligns LLMs using the DPO algorithm. The core innovation lies in its preference data construction strategy: it integrates AI-driven retrieval for factual grounding, enhancing knowledgeability and accuracy, and emphasizes the optimization of domain-specific CoT, treating the reasoning process itself as a key preference dimension. A multi-stage, AI-driven refinement pipeline cost-effectively generates these preference pairs. Experimental validation in Traditional Chinese Medicine (TCM) using Qwen3-1.7B as the base model demonstrates that RACE-Align significantly outperforms the original base model and a model fine-tuned only with Supervised Fine-Tuning (SFT). Improvements were observed across multiple dimensions, including answer accuracy, information richness, application of TCM thinking patterns, logicality and depth of reasoning, and interpretability. These findings suggest RACE-Align offers an effective pathway to enhance LLMs' knowledge application, reasoning reliability, and process transparency in complex vertical domains.
\end{abstract}


\section{Introduction}

In recent years, Large Language Models (LLMs) have developed rapidly, demonstrating formidable capabilities in numerous tasks such as natural language understanding and generation \cite{annepaka2024large,karamolegkou2025nlp}, heralding their extensive application potential across various industries. However, in vertical domains such as healthcare, law, and scientific research, which demand high levels of knowledge accuracy, logical rigor, and process transparency, LLMs still face challenges: ensuring factual accuracy, keeping pace with domain knowledge updates, generating reliable reasoning that conforms to professional paradigms, and providing interpretability \cite{busch2025current, matarazzo2025survey}.

\begin{figure*}[h]
    \centering
    \includegraphics[width=\textwidth]{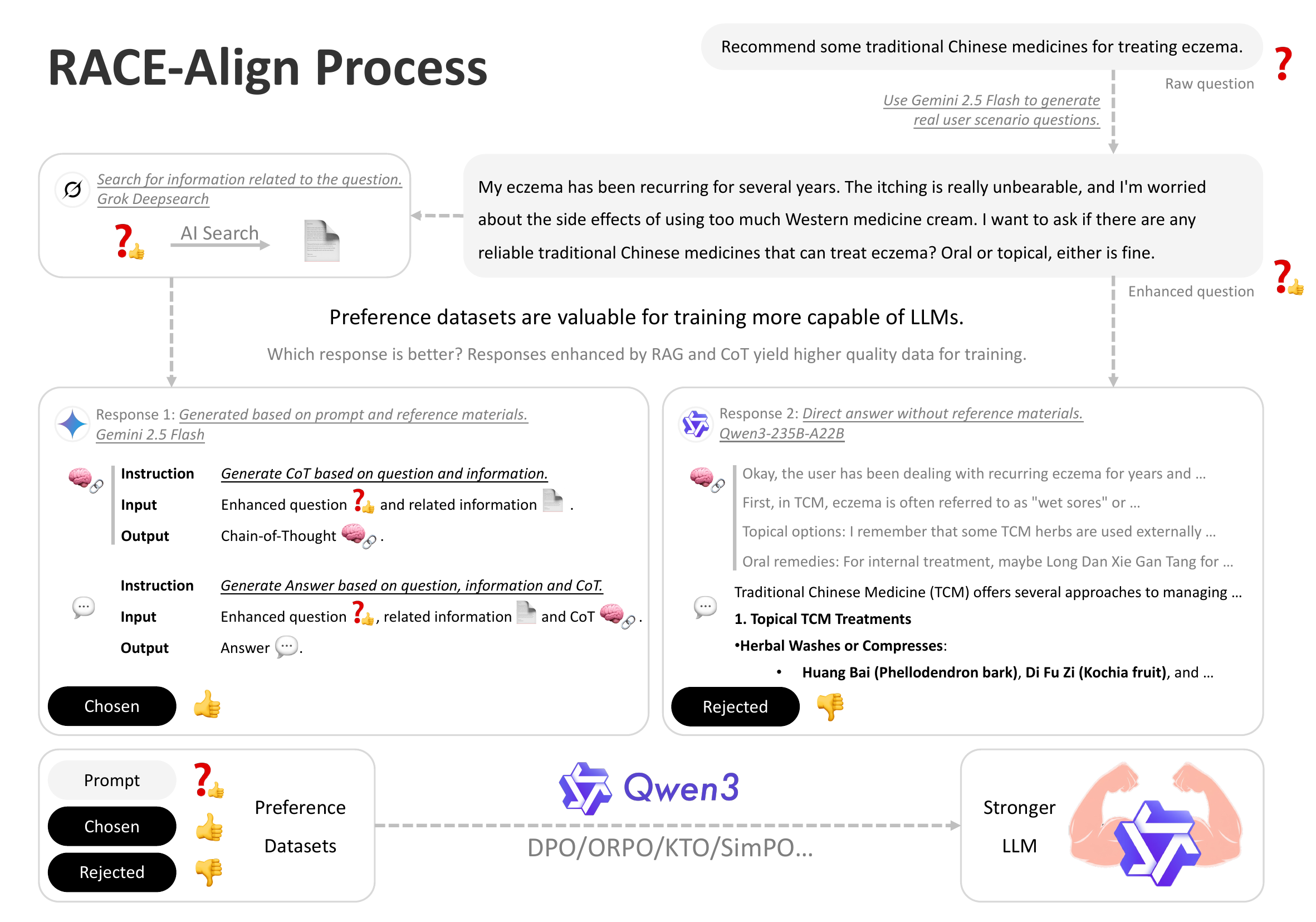}
    \caption{The RACE-Align framework flowchart. The process begins with an original user question, which is transformed into an enhanced question through Artificial Intelligence (AI) augmentation (e.g., using Gemini 2.5 Flash to generate more specific, contextualized questions). Subsequently, retrieval augmentation (e.g., Grok Deepsearch) is employed to acquire external knowledge relevant to the enhanced question. Based on the enhanced question and retrieved knowledge, a preferred answer (Chosen) is generated through a two-stage process: first, an instruction model (e.g., Gemini 2.5 Flash) generates a preliminary analysis containing an explicit Chain-of-Thought (CoT); then, based on the question, retrieved knowledge, and the generated CoT, the instruction model produces the final answer. Concurrently, another model (e.g., Qwen3-235B-A22B) is used to generate a rejected answer (Rejected) that inadequately utilizes external references or employs a more direct reasoning process. The resulting preference data pairs, comprising (prompt, chosen answer, rejected answer), are used to align a base Large Language Model (LLM) (e.g., Qwen3-1.7B) via algorithms like Direct Preference Optimization (DPO), aiming to produce a model that exhibits stronger performance, more reliable reasoning, and greater process transparency in specific vertical domains.}
    \label{fig:race_align_process}
\end{figure*}

Existing LLM alignment techniques, such as Reinforcement Learning from Human Feedback (RLHF) and its variant Direct Preference Optimization (DPO) \cite{rafailov2023direct}, have made significant contributions to enhancing model utility and harmlessness \cite{wang2024reinforcement}. However, these methods often treat the model as a black box, focusing on optimizing the final output while paying insufficient attention to the model's internal reasoning process. In vertical domains, users require not only correct answers but also an understanding of their derivation process, which is crucial for decision-making interpretability and reasoning transparency \cite{bilal2025llms,zhao2024explainability}. This demand has spurred the academic community to explore new preference alignment methods that consider both answer quality and reasoning process optimization. Some studies have begun to explore the integration of retrieval augmentation, Chain-of-Thought (CoT) reasoning, and preference learning \cite{dong2025understand, zhang2024chain, bi2024context, wu2024pa}, but how to systematically combine these elements, especially for explicitly optimizing the reasoning process itself, remains an area for exploration.

To address this, this paper proposes RACE-Align (Retrieval-Augmented and Chain-of-Thought Enhanced Alignment), a novel preference alignment framework, the detailed process of which is illustrated in Figure~\ref{fig:race_align_process}. The core of this framework lies in constructing a novel binary preference dataset that incorporates external knowledge utilization and explicit reasoning process quality as key dimensions for preference learning, and then aligning the LLM using DPO. The RACE-Align framework innovatively integrates Retrieval-Augmented Generation (RAG), explicit CoT Optimization, and DPO.

The main contributions include:
\begin{itemize}
\item We propose the RACE-Align framework, which innovatively integrates RAG, explicit CoT Optimization, and DPO. Although existing works have separately explored RAG with DPO (e.g., \cite{wu2024pa, zhang2025knowpo, dong2025understand}) or CoT with DPO (e.g., \cite{zhang2024chain, pang2024iterative, wang2024self}), and while \cite{bi2024context} involves all three, its primary goal is to enhance contextual faithfulness. In contrast, RACE-Align focuses more on treating the reasoning process itself as a core optimization dimension, combining retrieved knowledge to improve the knowledgeability, reasoning ability, and interpretability of applications in vertical domains.

\item We design and implement a multi-stage, AI-driven preference data generation pipeline to efficiently construct preference pairs containing external knowledge support and detailed reasoning chains. By integrating AI-powered web search to introduce external knowledge \cite{Yadav2024External,cheng2025survey}, we ensure the factuality and information richness of preferred samples. Simultaneously, we guide the model to generate and optimize domain-specific thought chains (CoT) \cite{wei2022chain}, making the reasoning process one of the core learning objectives.

\item We plan comprehensive experiments to validate the effectiveness of RACE-Align in enhancing LLM performance in vertical domains (answer accuracy, information content) and interpretability (reasoning logic rationality, clarity).
\end{itemize}
The remainder of this paper is organized as follows: Section 2 reviews related work. Section 3 elaborates on the RACE-Align method. Section 4 describes the experimental setup. Section 5 presents the experimental results. Section 6 discusses the findings. Section 7 concludes the paper.

\section{Related Work}

To better position the RACE-Align method, this section reviews research progress in three main related areas: large language model alignment, retrieval-augmented generation, and reasoning process optimization.

Regarding alignment techniques for LLMs, Supervised Fine-tuning (SFT) is the most fundamental method \cite{tie2025survey,guan2025survey}. However, SFT struggles to fully capture complex human preferences. To address this, RLHF learns human preferences by training a reward model, which in turn optimizes the LLM policy \cite{zhang2025mm,noukhovitch2024asynchronous,liang2025rlhs}. RLHF has achieved great success in improving model performance but also faces issues such as training instability and implementation complexity \cite{hu2025reinforce++,xie2024exploratory}. DPO \cite{rafailov2023direct} has emerged as a simpler and more effective alternative. It bypasses the explicit reward model training and directly optimizes the LLM policy from preference data, demonstrating better stability and efficiency. DPO has been widely applied to various alignment tasks, including enhancing CoT reasoning capabilities \cite{wang2024self, pang2024iterative, zhang2024chain} and optimizing generators in RAG systems \cite{wu2024pa, dong2025understand, xu2024retrievaldpo}. Furthermore, automated preference data generation has become an important research direction. Works such as Safer-Instruct \cite{shi2023safer}, Selfee \cite{dong2024self}, SynPO \cite{shahzad2025comprehensive}, and Constitutional AI \cite{bai2022constitutional} have explored how to leverage AI's own capabilities to construct preference data at scale, providing valuable insights for our design of an AI-driven multi-stage data generation pipeline.

RAG \cite{ZhangWang2025RAG, gao2023retrieval, liu2025improving} is a technique effective in mitigating LLM limitations such as knowledge cutoffs, outdatedness, and hallucinations. Its core idea is to retrieve relevant information from external knowledge bases as context before the model generates a response, thereby producing more accurate and factual content. RAG has been widely used in domain-specific question-answering (QA) systems \cite{WangShi2025KGRA_EN,FengHe2025RGRKBQA, han2024retrieval}. Recently, researchers have begun to explore applying RAG principles to preference data construction or alignment processes. For example, \cite{wu2024pa} constructed multi-perspective preference data and combined SFT and DPO to optimize generators in RAG scenarios. \cite{zhang2025knowpo} addressed knowledge conflicts in RAG by constructing simulated conflict datasets and using DPO to enable models to adaptively select knowledge. \cite{dong2025understand} proposed a dual preference alignment framework to jointly optimize retrievers and generators. \cite{xu2024retrievaldpo} utilized retrieval to form preference pairs from unpaired data. SWiRL \cite{goldie2025synthetic} combined search with fine-grained feedback on multi-step reasoning trajectories. Notably, \cite{bi2024context}, in RAG scenarios, constructed "faithful" versus "stubborn" answer pairs and introduced multi-hop knowledge links to build reasoning chains, using DPO to train models to be more faithful to the retrieved context. These works indicate that incorporating external knowledge into preference alignment is a promising direction for improving model performance. RACE-Align builds upon this by not only using retrieved knowledge to enhance the quality of the final answer but also emphasizing the explicit integration of this knowledge into the construction and optimization of reasoning chains in preferred samples.

Optimizing the reasoning process of LLMs, particularly through the guidance of CoT \cite{zhang2025enhancing,kim2024transformers,xu2025chain}, has become a key technique for enhancing model performance on complex tasks. CoT significantly improves logical reasoning capabilities by prompting LLMs to generate step-by-step intermediate reasoning steps \cite{qu2025tool}. How to further enhance LLM reasoning abilities through alignment techniques has also received widespread attention. For instance, Lightman \cite{wang2023making} explored using CoT reasoning to enhance alignment effects. More directly, PORT \cite{lahlou2024port} focused on applying preference optimization to the CoT reasoning steps themselves. Similarly, CPO \cite{zhang2024chain} created preference data using tree-like thought structures (e.g., Tree-of-Thoughts \cite{xia2024beyond}) and used DPO to optimize chain-like reasoning. Additionally, \cite{pang2024iterative} iteratively optimized reasoning chains for mathematical reasoning tasks; \cite{wang2024self} combined self-training with DPO fine-tuning; Preference Tree Optimization (TPO) \cite{baruchpreference} treated multi-branch reasoning structures as complete preference lists for learning. These studies all demonstrate that CoT preference optimization can significantly improve model reasoning capabilities. However, they are generally studied in settings without external retrieval and pay less attention to how retrieved knowledge can be integrated into the preference learning of reasoning chains. The uniqueness of RACE-Align lies in the fact that the CoTs it constructs and optimizes are based on externally retrieved knowledge, making the reasoning process not only more logically rigorous but also more factually grounded and domain-specific in content.

Finally, the exploration of LLM applications in specific vertical domains (such as Traditional Chinese Medicine (TCM) \cite{ren2025large, dai2024tcmchat, wei2024biancang, zhang2024qibo}, law \cite{schwarcz2025ai}, and finance \cite{yeo2025comprehensive}) is increasing. These applications commonly face a reliance on deep professional knowledge, high demands for logical rigor in reasoning, and an urgent user need for interpretability. These challenges highlight the necessity of alignment methods like RACE-Align, which aim to simultaneously enhance knowledge accuracy, reasoning reliability, and process interpretability. In terms of preference alignment research in vertical domains, KnowPAT \cite{tu2024findings} targets enterprise product QA scenarios, retrieving knowledge from knowledge graphs and constructing style and knowledge preference sets for alignment. ALFA \cite{li2025aligning} focuses on interactive diagnostic tasks for clinical reasoning, aligning fine-grained attributes. These works demonstrate approaches for improving LLM performance in vertical domains through preference learning, but differ from RACE-Align in their combination of methods (particularly RAG+CoT+DPO) and optimization objectives.

In summary, existing research has made significant progress in RAG, CoT optimization, and preference data generation. However, there is a lack of comprehensive research that systematically integrates retrieval augmentation (providing a knowledge base) with explicit reasoning chain optimization (enhancing interpretability and logicality, and treating the reasoning process itself as a training signal) into the construction of preference datasets, all unified under a DPO alignment framework to meet the demands of complex vertical domains. Although works like \cite{bi2024context} have involved RAG and reasoning chains, their focus is on contextual faithfulness, which differs from RACE-Align's emphasis on direct alignment optimization of reasoning steps. Other similar works often cover only one or two of these elements \cite{wu2024pa, zhang2024chain}. Therefore, from both methodological and conceptual perspectives, RACE-Align possesses uniqueness, does not entirely overlap with existing research, and instead integrates and extends multiple cutting-edge ideas, aiming to achieve synergistic improvements in knowledgeability, reasoning, and interpretability.

\section{RACE-Align Method}

This section will elaborate on the RACE-Align framework, first introducing its core design philosophy, then detailing its multi-stage preference data generation pipeline, and finally explaining how the generated preference data is used for model alignment.

\subsection{Core Idea}

The primary objective of the RACE-Align framework is to generate a high-quality binary preference dataset. This dataset, denoted as $D$, is structured as follows:
\begin{equation} \label{eq:preference_dataset_D}
D = \{(q_i, y_w^i, y_l^i)\}_{i=1}^N
\end{equation}
In this dataset, each instance $i$ consists of an input question $q_i$, a corresponding preferred response $y_w^i$, and a rejected response $y_l^i$. Unlike traditional preference data, $y_w^i$ and $y_l^i$ here not only contain the final textual answer but also incorporate a composite structure of explicit reasoning processes (i.e., CoT).

Two key assumptions underpin the design of RACE-Align: First, in vertical domains, high-quality answers must not only be accurate and informative, but their reasoning processes must also conform to domain-specific professional logic, be evidence-based, and be easily understood and verifiable by experts. Second, by enabling LLMs to explicitly learn preferred reasoning patterns and how to effectively utilize external knowledge during the alignment process, their interpretability, reliability, and overall performance in that domain can be significantly enhanced.

Based on these assumptions, the RACE-Align method is built upon two core pillars: The first is \textbf{Retrieval-Augmented}, where, during the generation of preferred samples, relevant knowledge is proactively retrieved from external sources (such as professional databases, real-time web information). This retrieved information serves as the factual basis for generating high-quality reasoning chains and final answers, thereby ensuring the accuracy, depth, and timeliness of the preferred samples. The second is \textbf{Chain-of-Thought Enhanced}, which treats detailed, structured CoT as an indispensable component of both preferred and rejected samples. By including these reasoning trajectories in the preference data and implicitly learning preferences over them during DPO alignment, the aim is to directly optimize the model's reasoning capabilities and logical expression.

\subsection{Preference Data Generation Pipeline}

The preference data generation for RACE-Align is a multi-stage, AI-driven refinement pipeline designed to systematically and scalably construct high-quality preference pairs that align with the aforementioned core ideas.

The first stage is question preparation and enhancement. This stage takes a set of raw domain questions $Q_{raw}$ as input, with the goal of transforming them into more natural, contextually richer enhanced questions with clear retrieval intent. Specifically, a pre-trained LLM (e.g., Gemini 2.5 Flash) is first used, guided by pre-defined templates covering various rewriting dimensions (such as sentence transformation, colloquialization, user perspective simulation, etc.), to rewrite an original question $q_{raw} \in Q_{raw}$ into an enhanced question $q_{enhanced}$, while ensuring the core query intent of the original question is fully preserved. Finally, the generated data entries $(q_{raw}^i, q_{enhanced}^i)$ are consolidated and assigned unique IDs, forming a structured enhanced question dataset $D_{enhanced}$:
\begin{equation} \label{eq:dataset_enhanced}
D_{enhanced} = \{(id_i, q_{raw}^i, q_{enhanced}^i)\}_{i=1}^N
\end{equation}
where $id_i$ is a unique identifier for the $i$-th sample.

The second stage focuses on the generation of rejected samples. This stage takes the enhanced question $q_{enhanced}^i$ from the previous stage as input. The objective is to generate a preliminary answer for each question, which may have deficiencies (such as knowledge errors, logical leaps), along with its corresponding thought process, to serve as the rejected party $y_l^i$ in the preference pair. This is achieved by utilizing a capable LLM (e.g., Qwen3-235B-A22B) to directly respond to $q_{enhanced}^i$. By appending specific instructions (such as the \texttt{/think} command commonly used in Qwen series models), the model is guided to output its internal raw thought process $reasoning_l^i$ while generating the final answer $answer_l^i$. These two parts combined define the rejected sample $y_l^i$:
\begin{equation} \label{eq:rejected_sample_structure}
y_l^i = (reasoning_l^i, answer_l^i)
\end{equation}
After this stage, the dataset is updated to $D_{rejected}$:
\begin{equation} \label{eq:dataset_rejected}
D_{rejected} = \{(id_i, q_{raw}^i, q_{enhanced}^i, y_l^i)\}_{i=1}^N
\end{equation}

The third stage involves external knowledge integration (RAG). This stage uses the enhanced question $q_{enhanced}^i$ from $D_{rejected}$ to guide retrieval. The goal is to retrieve relevant, high-quality professional knowledge from external sources for each question, denoted as $RAG\_content^i$. Specifically, AI tools or models with strong web search and information integration capabilities (such as Grok DeepSearch, or a combination of search engine Application Programming Interfaces (APIs) and LLMs for summary extraction) are employed for deep knowledge retrieval related to $q_{enhanced}^i$. After obtaining the raw results, preprocessing and filtering are performed, such as removing irrelevant markers, consolidating information, and ensuring content relevance and information density, to finally obtain the integrated external knowledge $RAG\_content^i$. The dataset is then updated to $D_{RAG}$:
\begin{equation} \label{eq:dataset_rag}
D_{RAG} = \{(id_i, q_{raw}^i, q_{enhanced}^i, y_l^i, RAG\_content^i)\}_{i=1}^N
\end{equation}

The fourth stage, preferred sample generation, is a critical part of the pipeline, aimed at producing high-quality preferred samples $y_w^i$. It consists of two closely related sub-steps. First, in sub-step 4.1, the preferred reasoning chain (CoT) is constructed. This step takes the enhanced question $q_{enhanced}^i$ and the corresponding external knowledge $RAG\_content^i$ from $D_{RAG}$ as input. The objective is to generate a detailed, logically rigorous CoT $reasoning_w^i$ that effectively integrates RAG knowledge. An LLM with strong reasoning and generation capabilities (e.g., Gemini 2.5 Flash) is utilized, guided by a carefully designed prompt. This prompt instructs the model to understand the question, apply domain theories, critically and implicitly incorporate $RAG\_content^i$ information, form a coherent and clear thought process, and potentially provide preliminary conclusions. The generated CoT is typically output in a structured format (e.g., JSON array) and subsequently merged into a single text string $reasoning_w^i$. Next, in sub-step 4.2, the preferred formal answer is generated. This step takes the original question $q_{raw}^i$ from $D_{RAG}$ (the original question is chosen so that the final answer more directly corresponds to the user's initial query), the relevant external knowledge $RAG\_content^i$, and the preferred reasoning chain $reasoning_w^i$ generated in the previous step as input. The goal is to generate a user-oriented, well-formatted (e.g., Markdown), informative, and accurate preferred answer $answer_w^i$. Again, an LLM (e.g., Gemini 2.5 Flash) and a new prompt are used, instructing the model to strictly follow the logical framework of $reasoning_w^i$, appropriately utilize $RAG\_content^i$ to enrich the content, use clear and professional language, present in the required format, and ensure the content is free of private or inappropriate information. After completing these two sub-steps, the preferred sample $y_w^i$ is obtained:
\begin{equation} \label{eq:preferred_sample_structure}
y_w^i = (reasoning_w^i, answer_w^i)
\end{equation}
The final preference dataset can be represented as $D_{final}$:
\begin{equation} \label{eq:dataset_final}
D_{final} = \{(id_i, q_{raw}^i, q_{enhanced}^i, y_l^i, y_w^i, RAG\_content^i)\}_{i=1}^N
\end{equation}
During DPO training, the question (typically $q_{enhanced}^i$ to leverage its context), the rejected sample $y_l^i$, and the preferred sample $y_w^i$ are primarily used.

Finally, the fifth stage is DPO dataset formatting. This stage converts the $D_{final}$ dataset into the input format required by a specific DPO training framework (e.g., Qwen3's training framework). The goal is to generate a structure for each data entry containing three fields: \texttt{prompt}, \texttt{chosen}, and \texttt{rejected}. Specifically, the \texttt{prompt} field usually contains the user's question (e.g., using $q_{enhanced}^i$ so the model can utilize its enhanced context) and may include special instructions (e.g., \texttt{/think}) to prompt the model to activate its thinking mode. The \texttt{chosen} field contains the preferred response, composed of the preferred thought process $reasoning_w^i$ (embedded with specific tags like \texttt{\textless{}think\textgreater{}}$\dots$\texttt{\textless{}/think\textgreater{}}) and the preferred final answer $answer_w^i$. The \texttt{rejected} field has a similar structure, containing the rejected thought process $reasoning_l^i$ (also embedded with \texttt{\textless{}think\textgreater{}} tags) and the rejected final answer $answer_l^i$. The output of this stage is a dataset $D_{DPO}$ that can be directly used for DPO training:
\begin{equation} \label{eq:dataset_dpo}
D_{DPO} = \{(\text{prompt}_i, \text{chosen}_i, \text{rejected}_i)\}_{i=1}^N
\end{equation}

\subsection{Model Alignment}

After obtaining the formatted DPO preference dataset $D_{DPO}$, we employ the DPO algorithm \cite{rafailov2023direct} to align the pre-trained LLM. The core idea of DPO is to directly use preference pairs $(q, y_w, y_l)$ to optimize the language model's policy $\pi_\theta$, making it more inclined to generate the preferred response $y_w$ over the rejected response $y_l$. The DPO objective function can be expressed as:
\begin{equation} \label{eq:dpo_loss}
\mathcal{L}_{\mathrm{DPO}}(\pi_\theta; \pi_{\mathrm{ref}}) = -\mathbb{E}_{(q, y_w, y_l) \sim D_{\mathrm{DPO}}} \left[ \log \sigma \left( \beta \left( \log \frac{\pi_\theta(y_w|q)}{\pi_{\mathrm{ref}}(y_w|q)} - \log \frac{\pi_\theta(y_l|q)}{\pi_{\mathrm{ref}}(y_l|q)} \right) \right) \right]
\end{equation}
where $\pi_\theta$ is the language model policy we wish to align through optimization; $\pi_{\mathrm{ref}}$ is a reference policy, which is the initial version of $\pi_\theta$ before DPO, used for regularization to prevent $\pi_\theta$ from deviating too far during optimization; $q$ is the input question (i.e., the \texttt{prompt} in the DPO dataset); $y_w$ is the preferred response (i.e., the \texttt{chosen} in the DPO dataset, containing the reasoning chain and final answer); $y_l$ is the rejected response (i.e., the \texttt{rejected} in the DPO dataset, also containing the reasoning chain and final answer); $\sigma(\cdot)$ is the Sigmoid function; and $\beta$ is a hyperparameter that controls the weight of the KL divergence between the reference policy and the optimized policy, effectively adjusting the model's sensitivity to the preference signal.

By minimizing this loss function (as shown in Equation~\eqref{eq:dpo_loss}), the model $\pi_\theta$ is trained to increase the likelihood ratio of generating $y_w$ relative to $y_l$ (compared to the likelihood ratio of the reference model $\pi_{\mathrm{ref}}$). Importantly, under the RACE-Align framework, because both $y_w$ and $y_l$ embed explicit reasoning processes (CoT), the model implicitly learns preferences for reasoning processes while learning preferences for final answers. This means the model is not only encouraged to generate better answers but also to adopt better ways of thinking to arrive at those answers.

\section{Experimental Setup}

To validate the effectiveness of the proposed RACE-Align method, we designed a meticulous experimental procedure. This section will detail the dataset construction process, the chosen base model, the setup of comparative models, training specifics, and the evaluation scheme.

\subsection{Dataset Construction}

The experimental data for this study originates from the publicly available TCM domain QA dataset, ShenNong\_TCM\_Dataset \cite{zhu2023ChatMed}. We randomly selected 5000 original question-answering data entries from it as a foundation. Subsequently, following the RACE-Align methodology detailed in Section 3 of this paper, we systematically rewrote and enhanced these 5000 original data entries and constructed a binary preference dataset structured as (question, preferred answer and reasoning, rejected answer and reasoning). A core feature of this dataset is that the preferred samples $y_w$ not only contain retrieval-augmented, factually accurate, and information-rich answers but also integrate exhaustive, domain-logically consistent explicit CoT. Correspondingly, the rejected samples $y_l$ also include their reasoning processes and final answers for comparative learning. During the construction process, we utilized advanced AI models such as Gemini 2.5 Flash and Qwen3-235B-A22B to assist with question enhancement, preliminary answer generation, external knowledge integration, and the construction and refinement of preferred reasoning chains and answers. From the finally generated 5000 high-quality preference data pairs, we selected the first 4900 as the training set and the latter 100 as the validation set, used to evaluate the model's generalization ability on unseen data.

\subsection{Base Model and Comparative Models}

The experiment employed the Qwen3-1.7B model, recently open-sourced by the Qwen team, as the base LLM. This model was chosen for its strong general language understanding and generation capabilities, although its expertise in specific vertical domains (like TCM) and its domain-specific reasoning paradigms require dedicated optimization. We anticipated this model to perform well in hallucination control but primarily rely on conventional CoT for analysis, lacking the application of TCM-specific thinking patterns.

To comprehensively evaluate the effectiveness of the RACE-Align method, we designed the following three models for comparison:
The first is the original Qwen3-1.7B base model without any additional training (hereinafter referred to as "Base Model").
The second is a model fine-tuned using only the preferred answer portion ($y_w$) of our constructed preference data via SFT (hereinafter referred to as "SFT Model"). Specifically, the SFT training data format is $(q_{enhanced}, y_w)$, where $q_{enhanced}$ is the enhanced question, and $y_w$ contains the preferred reasoning chain and preferred answer.
The third is a model trained using the RACE-Align method proposed in this paper, utilizing the complete preference dataset (containing $(q_{enhanced}, y_w, y_l)$) to perform DPO on Qwen3-1.7B (hereinafter referred to as "RACE-Align DPO Model").

\subsection{Training Details}

All model training and fine-tuning were conducted in a unified computational environment to ensure the comparability of experimental results.

For the SFT Model, we adopted a standard supervised learning paradigm, using a cross-entropy loss function to optimize model parameters. Hyperparameters such as learning rate, batch size, and training epochs were selected through preliminary experiments on the validation set. For the RACE-Align DPO Model, we followed the DPO algorithm described in Section 3.3, with the loss function as shown in Equation~\eqref{eq:dpo_loss}. The key hyperparameter $\beta$ was adjusted based on literature recommendations and preliminary experiments to balance preference learning strength and divergence from the reference model. In DPO training, the reference model $\pi_{\mathrm{ref}}$ was the model trained via SFT, a common practice aimed at stabilizing the DPO training process and retaining useful knowledge learned during the SFT phase. The training process utilized the Hugging Face Transformers library and PyTorch framework, and was executed on appropriate GPU hardware.

\subsection{Evaluation Methods}

To comprehensively measure the performance of different models on vertical domain QA tasks, we conducted evaluations from two dimensions: answer quality and reasoning process, using a total of 100 sets of evaluation samples. The evaluation method combined automated metrics and human assessment.

For automated evaluation of answer quality, we employed the advanced LLM Gemini 2.5 Pro as an evaluator. Gemini 2.5 Pro scored the answers generated by each model across several common LLM evaluation dimensions, such as accuracy, relevance, and information richness. Additionally, traditional automated Natural Language Generation (NLG) metrics like Recall-Oriented Understudy for Gisting Evaluation (ROUGE) (ROUGE-1, ROUGE-2, ROUGE-L) \cite{lin2004rouge} and Bilingual Evaluation Understudy (BLEU) \cite{papineni2002bleu} were also used to assess the similarity in content coverage between generated answers and reference answers.

Human evaluation is a core component of this experiment, designed to deeply examine the model's performance in applying domain-specific knowledge and reasoning logic. We recruited 5 volunteers with a background in TCM as evaluators. The evaluators conducted a blind review of the complete responses (including reasoning chains and final answers) generated by each model on the validation set (100 sample sets). The evaluation dimensions primarily included:
1) Answer Accuracy and Factuality: Whether the content conforms to TCM professional knowledge, whether there are factual errors or hallucinations, and whether (simulated) retrieved knowledge is effectively utilized.
2) Application of TCM Thinking Patterns: Whether the model can use TCM-specific theories and logic (such as Zang-Fu organ pattern differentiation, Five Elements theory, etc.) for analysis and reasoning.
3) Logicality and Depth of Reasoning: Whether the thought chain is clear, logically coherent, with reasonable steps, and whether the reasoning process has sufficient depth and insight.
4) Information Richness and Insightfulness: Whether the answer is comprehensive, in-depth, and provides valuable information and unique insights beyond simple factual statements.
5) Interpretability and Transparency: Whether the reasoning process is easy to understand, whether it can clearly show the model's thought path to the answer, enhancing user trust in the model's output.
Evaluators scored each dimension (e.g., using a 1-5 point Likert scale) and provided qualitative feedback.

In addition to quantitative evaluation, we also conducted detailed case studies, selecting representative questions to compare the answers and reasoning processes generated by different models, to more intuitively demonstrate the specific advantages and potential shortcomings of the RACE-Align method in enhancing knowledgeability, reasoning, and interpretability.

\section{Experimental Results}

This section will present the experimental data obtained through automated metrics and human evaluation in detail, to compare the specific performance of different models across various evaluation dimensions.

\subsection{Automated Evaluation Results}

Automated evaluation was primarily conducted using the Gemini 2.5 Pro model and traditional NLG evaluation metrics (such as ROUGE, BLEU). Table~\ref{tab:auto_eval} summarizes the main performance of each model in the automated evaluation.

\begin{table}[H] % Using H for "here definitely" from float package
\centering
\caption{Automated Evaluation Results. Metrics include Gemini 2.5 Pro scores (out of 10), ROUGE-L, and BLEU-4.}
\label{tab:auto_eval}
\begin{tabular}{@{}lccccc@{}}
\toprule
Model & \shortstack{Information\\Richness} & \shortstack{Relevance} & \shortstack{Accuracy} & \shortstack{ROUGE-L} & \shortstack{BLEU-4} \\
\midrule
Base Model (Qwen3-1.7B) & 7.0 & 8.0 & \textbf{8.5} & 0.55 & 0.30 \\
SFT Model & 7.8 & 7.5 & 7.0 & 0.58 & 0.32 \\
RACE-Align DPO Model & \textbf{8.5} & \textbf{8.8} & 8.0 & \textbf{0.62} & \textbf{0.35} \\
\bottomrule
\end{tabular}
\end{table}

As seen in Table~\ref{tab:auto_eval}, the RACE-Align DPO model performed best on metrics measuring content quality and generation effectiveness, such as information richness, relevance, ROUGE-L, and BLEU-4. The Base Model (Qwen3-1.7B) scored highest in accuracy (primarily reflecting low hallucination), which is consistent with its characteristics as a powerful general-purpose model. The SFT Model's performance was intermediate across various metrics but was lower in accuracy than the RACE-Align DPO model. These data preliminarily indicate the potential of the RACE-Align framework in enhancing content quality and domain adaptability, while also maintaining a good level of accuracy, superior to the simple SFT method.

\subsection{Human Evaluation Results}

The blind human evaluation conducted by 5 volunteers with a background in TCM provided deeper insights. Table~\ref{tab:human_eval} summarizes the main results of the human evaluation, with all scores based on a 1-5 point Likert scale, where 5 is the highest.

\begin{table}[H] % Using H for "here definitely" from float package
\centering
\caption{Human Evaluation Results (Average score, out of 5).}
\label{tab:human_eval}
\resizebox{\textwidth}{!}{%
\begin{tabular}{@{}lccccc@{}}
\toprule
Model & \shortstack{Answer Accuracy\\\& Factuality} & \shortstack{TCM Thinking\\Pattern Application} & \shortstack{Reasoning Logicality\\\& Depth} & \shortstack{Information Richness\\\& Insightfulness} & \shortstack{Interpretability\\\& Transparency} \\
\midrule
Base Model (Qwen3-1.7B) & \textbf{4.2} & 1.5 & 3.5 & 3.0 & 3.2 \\
SFT Model & 3.0 & 3.8 & 3.6 & 3.9 & 3.7 \\
RACE-Align DPO Model & 3.8 & \textbf{4.5} & \textbf{4.4} & \textbf{4.6} & \textbf{4.3} \\
\bottomrule
\end{tabular}%
}
\end{table}

From the human evaluation results in Table~\ref{tab:human_eval}, the RACE-Align DPO model performed optimally across most key dimensions. Particularly in the application of TCM thinking patterns, logicality and depth of reasoning, information richness and insightfulness, and interpretability and transparency, the RACE-Align DPO model received the highest average scores. This indicates that the model can more effectively apply TCM theories for in-depth analysis and present its thought processes and insightful answers in a clear, understandable manner. The Base Model (Qwen3-1.7B) performed best in answer accuracy and factuality, primarily due to its strong pre-training foundation and good hallucination control. The SFT Model's performance was intermediate between the two; although it could apply some TCM thinking, it lagged behind the RACE-Align DPO model in depth and accuracy, and was less factual than the base model. These results further highlight the significant effectiveness of the RACE-Align method in enhancing the model's domain expertise and reasoning capabilities.

\section{Discussion}

Synthesizing the automated and human evaluation results, RACE-Align demonstrates its effectiveness in enhancing LLM performance in vertical domains like TCM. The results highlight that while the Qwen3-1.7B base model is fluent and controls hallucinations, it lacks deep domain understanding and specific reasoning paradigms. SFT, while enabling some learning of domain terminology, offers limited improvement in depth and logical rigor, suggesting that supervised learning on preferred answers alone is insufficient for internalizing complex domain knowledge and reasoning.

In contrast, the RACE-Align DPO model, by deeply integrating RAG mechanisms and explicit CoT optimization, exhibits significant performance advantages. Training on preference data pairs incorporating external knowledge and detailed reasoning processes allows the model to achieve a more profound understanding of domain-specific analytical thinking and reasoning methods. This leads to answers with greater professional insight and information richness, alongside better factual accuracy compared to the SFT model. The core contribution lies in unifying effective external knowledge utilization with explicit internal reasoning process optimization under a DPO framework, thereby enhancing the model's intelligence, credibility, and transparency. The AI-driven data generation pipeline also shows promise for transferability to other complex domains.

The RACE-Align method achieves these benefits cost-effectively due to its AI-driven pipeline, which significantly improves efficiency over traditional manual annotation. Nevertheless, the reliance on RAG necessitates careful attention to the quality of retrieved external knowledge. Furthermore, evaluating model reasoning processes and interpretability remains challenging; existing automated metrics often struggle with the nuances of logical reasoning, while human evaluation, though more reliable, is resource-intensive. This study, as a preliminary exploration, also faced limitations such as a restricted number of evaluators.

Looking ahead, future work should focus on several areas. These include continuously optimizing the efficiency and cost-control of preference data generation, for instance, through more intelligent sampling or by using more cost-effective LLMs for auxiliary tasks. Exploring dynamic adjustment of RAG and CoT strategies based on question complexity could lead to finer-grained optimization. Additionally, deeply integrating RACE-Align principles with other advanced alignment techniques (e.g., red teaming) or AI safety technologies (e.g., Constitutional AI) could further enhance model robustness. Extensively validating RACE-Align across a wider range of vertical domains is crucial to assess its universality, as is the continued investment in developing more refined and objective automated evaluation metrics for reasoning processes.

\section{Conclusion}

This paper proposed and validated a novel LLM preference alignment framework named RACE-Align, designed to address the core challenges of knowledge accuracy, reasoning logicality, and process interpretability faced by LLMs in specific vertical domain applications. The core innovation of RACE-Align lies in the deep integration of RAG and explicit CoT optimization into the DPO alignment process. We designed and implemented a multi-stage, AI-driven preference data generation pipeline capable of systematically and cost-effectively constructing binary preference data pairs that include external knowledge support and detailed reasoning steps. Experimental results demonstrate that, compared to the base model and traditional SFT methods, the model aligned with RACE-Align (using Qwen3-1.7B as an example, validated in the TCM domain) significantly outperforms the original base model and the SFT-only model across multiple dimensions, including answer knowledgeability, information richness, application of TCM thinking patterns, logicality and depth of reasoning, and interpretability. These findings indicate that RACE-Align provides an effective and novel pathway for enhancing LLMs' knowledge application capabilities, reasoning reliability, and process transparency in complex vertical domains, potentially promoting their successful application in more demanding scenarios.

%Bibliography
\bibliographystyle{unsrt}
\bibliography{references}

\begin{thebibliography}{10}

\bibitem{annepaka2024large}
Yadagiri Annepaka and Partha Pakray.
\newblock Large language models: A survey of their development, capabilities, and applications.
\newblock {\em Knowledge and Information Systems}, pages 1--56, 2024.

\bibitem{karamolegkou2025nlp}
Antonia Karamolegkou, Angana Borah, Eunjung Cho, Sagnik~Ray Choudhury, Martina Galletti, Rajarshi Ghosh, Pranav Gupta, Oana Ignat, Priyanka Kargupta, Neema Kotonya, et~al.
\newblock Nlp for social good: A survey of challenges, opportunities, and responsible deployment.
\newblock {\em arXiv preprint arXiv:2505.22327}, 2025.

\bibitem{busch2025current}
Felix Busch, Lena Hoffmann, Christopher Rueger, Elon~HC van Dijk, Rawen Kader, Esteban Ortiz-Prado, Marcus~R Makowski, Luca Saba, Martin Hadamitzky, Jakob~Nikolas Kather, et~al.
\newblock Current applications and challenges in large language models for patient care: a systematic review.
\newblock {\em Communications Medicine}, 5(1):26, 2025.

\bibitem{matarazzo2025survey}
Andrea Matarazzo and Riccardo Torlone.
\newblock A survey on large language models with some insights on their capabilities and limitations.
\newblock {\em arXiv preprint arXiv:2501.04040}, 2025.

\bibitem{rafailov2023direct}
Rafael Rafailov, Archit Sharma, Eric Mitchell, Christopher~D Manning, Stefano Ermon, and Chelsea Finn.
\newblock Direct preference optimization: Your language model is secretly a reward model.
\newblock {\em Advances in Neural Information Processing Systems}, 36:53728--53741, 2023.

\bibitem{wang2024reinforcement}
Shuhe Wang, Shengyu Zhang, Jie Zhang, Runyi Hu, Xiaoya Li, Tianwei Zhang, Jiwei Li, Fei Wu, Guoyin Wang, and Eduard Hovy.
\newblock Reinforcement learning enhanced llms: A survey.
\newblock {\em arXiv preprint arXiv:2412.10400}, 2024.

\bibitem{bilal2025llms}
Ahsan Bilal, David Ebert, and Beiyu Lin.
\newblock Llms for explainable ai: A comprehensive survey.
\newblock {\em arXiv preprint arXiv:2504.00125}, 2025.

\bibitem{zhao2024explainability}
Haiyan Zhao, Hanjie Chen, Fan Yang, Ninghao Liu, Huiqi Deng, Hengyi Cai, Shuaiqiang Wang, Dawei Yin, and Mengnan Du.
\newblock Explainability for large language models: A survey.
\newblock {\em ACM Transactions on Intelligent Systems and Technology}, 15(2):1--38, 2024.

\bibitem{dong2025understand}
Guanting Dong, Yutao Zhu, Chenghao Zhang, Zechen Wang, Ji-Rong Wen, and Zhicheng Dou.
\newblock Understand what llm needs: Dual preference alignment for retrieval-augmented generation.
\newblock In {\em Proceedings of the ACM on Web Conference 2025}, pages 4206--4225, 2025.

\bibitem{zhang2024chain}
Xuan Zhang, Chao Du, Tianyu Pang, Qian Liu, Wei Gao, and Min Lin.
\newblock Chain of preference optimization: Improving chain-of-thought reasoning in llms.
\newblock {\em Advances in Neural Information Processing Systems}, 37:333--356, 2024.

\bibitem{bi2024context}
Baolong Bi, Shaohan Huang, Yiwei Wang, Tianchi Yang, Zihan Zhang, Haizhen Huang, Lingrui Mei, Junfeng Fang, Zehao Li, Furu Wei, et~al.
\newblock Context-dpo: Aligning language models for context-faithfulness.
\newblock {\em arXiv preprint arXiv:2412.15280}, 2024.

\bibitem{wu2024pa}
Jiayi Wu, Hengyi Cai, Lingyong Yan, Hao Sun, Xiang Li, Shuaiqiang Wang, Dawei Yin, and Ming Gao.
\newblock Pa-rag: Rag alignment via multi-perspective preference optimization.
\newblock {\em arXiv preprint arXiv:2412.14510}, 2024.

\bibitem{zhang2025knowpo}
Ruizhe Zhang, Yongxin Xu, Yuzhen Xiao, Runchuan Zhu, Xinke Jiang, Xu~Chu, Junfeng Zhao, and Yasha Wang.
\newblock Knowpo: Knowledge-aware preference optimization for controllable knowledge selection in retrieval-augmented language models.
\newblock In {\em Proceedings of the AAAI Conference on Artificial Intelligence}, volume~39, pages 25895--25903, 2025.

\bibitem{pang2024iterative}
Richard~Yuanzhe Pang, Weizhe Yuan, He~He, Kyunghyun Cho, Sainbayar Sukhbaatar, and Jason Weston.
\newblock Iterative reasoning preference optimization.
\newblock {\em Advances in Neural Information Processing Systems}, 37:116617--116637, 2024.

\bibitem{wang2024self}
Tianduo Wang, Shichen Li, and Wei Lu.
\newblock Self-training with direct preference optimization improves chain-of-thought reasoning.
\newblock {\em arXiv preprint arXiv:2407.18248}, 2024.

\bibitem{Yadav2024External}
Ishita Yadav, Simon Schindler, David Peters, and Roman Klinger.
\newblock External knowledge integration in large language models: A survey on methods, challenges, and future directions.
\newblock {\em arXiv preprint arXiv:2403.11181}, 2024.

\bibitem{cheng2025survey}
Mingyue Cheng, Yucong Luo, Jie Ouyang, Qi~Liu, Huijie Liu, Li~Li, Shuo Yu, Bohou Zhang, Jiawei Cao, Jie Ma, et~al.
\newblock A survey on knowledge-oriented retrieval-augmented generation.
\newblock {\em arXiv preprint arXiv:2503.10677}, 2025.

\bibitem{wei2022chain}
Jason Wei, Xuezhi Wang, Dale Schuurmans, Maarten Bosma, Fei Xia, Ed~Chi, Quoc~V Le, Denny Zhou, et~al.
\newblock Chain-of-thought prompting elicits reasoning in large language models.
\newblock {\em Advances in neural information processing systems}, 35:24824--24837, 2022.

\bibitem{tie2025survey}
Guiyao Tie, Zeli Zhao, Dingjie Song, Fuyang Wei, Rong Zhou, Yurou Dai, Wen Yin, Zhejian Yang, Jiangyue Yan, Yao Su, et~al.
\newblock A survey on post-training of large language models.
\newblock {\em arXiv preprint arXiv:2503.06072}, 2025.

\bibitem{guan2025survey}
Jian Guan, Junfei Wu, Jia-Nan Li, Chuanqi Cheng, and Wei Wu.
\newblock A survey on personalized alignment--the missing piece for large language models in real-world applications.
\newblock {\em arXiv preprint arXiv:2503.17003}, 2025.

\bibitem{zhang2025mm}
Yi-Fan Zhang, Tao Yu, Haochen Tian, Chaoyou Fu, Peiyan Li, Jianshu Zeng, Wulin Xie, Yang Shi, Huanyu Zhang, Junkang Wu, et~al.
\newblock Mm-rlhf: The next step forward in multimodal llm alignment.
\newblock {\em arXiv preprint arXiv:2502.10391}, 2025.

\bibitem{noukhovitch2024asynchronous}
Michael Noukhovitch, Shengyi Huang, Sophie Xhonneux, Arian Hosseini, Rishabh Agarwal, and Aaron Courville.
\newblock Asynchronous rlhf: Faster and more efficient off-policy rl for language models.
\newblock {\em arXiv preprint arXiv:2410.18252}, 2024.

\bibitem{liang2025rlhs}
Kaiqu Liang, Haimin Hu, Ryan Liu, Thomas~L Griffiths, and Jaime~Fern{\'a}ndez Fisac.
\newblock Rlhs: Mitigating misalignment in rlhf with hindsight simulation.
\newblock {\em arXiv preprint arXiv:2501.08617}, 2025.

\bibitem{hu2025reinforce++}
Jian Hu.
\newblock Reinforce++: A simple and efficient approach for aligning large language models.
\newblock {\em arXiv preprint arXiv:2501.03262}, 2025.

\bibitem{xie2024exploratory}
Tengyang Xie, Dylan~J Foster, Akshay Krishnamurthy, Corby Rosset, Ahmed Awadallah, and Alexander Rakhlin.
\newblock Exploratory preference optimization: Harnessing implicit q*-approximation for sample-efficient rlhf.
\newblock {\em arXiv preprint arXiv:2405.21046}, 2024.

\bibitem{xu2024retrievaldpo}
M.~Xu, Z.~Chen, Z.~Liu, T.~Zhao, and D.~Yu.
\newblock Retrieval-dpo: Retrieval-augmented preference optimization with non-paired preference data.
\newblock OpenReview, 2024.

\bibitem{shi2023safer}
Taiwei Shi, Kai Chen, and Jieyu Zhao.
\newblock Safer-instruct: Aligning language models with automated preference data.
\newblock {\em arXiv preprint arXiv:2311.08685}, 2023.

\bibitem{dong2024self}
Qingxiu Dong, Li~Dong, Xingxing Zhang, Zhifang Sui, and Furu Wei.
\newblock Self-boosting large language models with synthetic preference data.
\newblock {\em arXiv preprint arXiv:2410.06961}, 2024.

\bibitem{shahzad2025comprehensive}
Tariq Shahzad, Tehseen Mazhar, Muhammad~Usman Tariq, Wasim Ahmad, Khmaies Ouahada, and Habib Hamam.
\newblock A comprehensive review of large language models: issues and solutions in learning environments.
\newblock {\em Discover Sustainability}, 6(1):27, 2025.

\bibitem{bai2022constitutional}
Yuntao Bai, Saurav Kadavath, Sandipan Kundu, Amanda Askell, Jackson Kernion, Andy Jones, Anna Chen, Anna Goldie, Azalia Mirhoseini, Cameron McKinnon, et~al.
\newblock Constitutional ai: Harmlessness from ai feedback.
\newblock {\em arXiv preprint arXiv:2212.08073}, 2022.

\bibitem{ZhangWang2025RAG}
Y.~Zhang and J.~Wang.
\newblock Retrieval-augmented generation: Advances and challenges in 2025.
\newblock {\em Journal of Artificial Intelligence Research}, 62(3):456--478, 2025.

\bibitem{gao2023retrieval}
Yunfan Gao, Yun Xiong, Xinyu Gao, Kangxiang Jia, Jinliu Pan, Yuxi Bi, Yixin Dai, Jiawei Sun, Haofen Wang, and Haofen Wang.
\newblock Retrieval-augmented generation for large language models: A survey.
\newblock {\em arXiv preprint arXiv:2312.10997}, 2(1), 2023.

\bibitem{liu2025improving}
Siru Liu, Allison~B McCoy, and Adam Wright.
\newblock Improving large language model applications in biomedicine with retrieval-augmented generation: a systematic review, meta-analysis, and clinical development guidelines.
\newblock {\em Journal of the American Medical Informatics Association}, page ocaf008, 2025.

\bibitem{WangShi2025KGRA_EN}
H.~Wang and Y.~Shi.
\newblock A survey on knowledge graph and retrieval-augmented generation for enhanced language model reasoning.
\newblock {\em Academic Journal of Science and Technology}, 14(1):227--235, 2025.

\bibitem{FengHe2025RGRKBQA}
T.~Feng and L.~He.
\newblock Rgr-kbqa: Generating question answering logical forms using knowledge graph enhanced large language models.
\newblock In {\em Proceedings of the 31st International Conference on Computational Linguistics}. Association for Computational Linguistics, 2025.

\bibitem{han2024retrieval}
Haoyu Han, Yu~Wang, Harry Shomer, Kai Guo, Jiayuan Ding, Yongjia Lei, Mahantesh Halappanavar, Ryan~A Rossi, Subhabrata Mukherjee, Xianfeng Tang, et~al.
\newblock Retrieval-augmented generation with graphs (graphrag).
\newblock {\em arXiv preprint arXiv:2501.00309}, 2024.

\bibitem{goldie2025synthetic}
Anna Goldie, Azalia Mirhoseini, Hao Zhou, Irene Cai, and Christopher~D Manning.
\newblock Synthetic data generation \& multi-step rl for reasoning \& tool use.
\newblock {\em arXiv preprint arXiv:2504.04736}, 2025.

\bibitem{zhang2025enhancing}
Yufeng Zhang, Xuepeng Wang, Lingxiang Wu, and Jinqiao Wang.
\newblock Enhancing chain of thought prompting in large language models via reasoning patterns.
\newblock In {\em Proceedings of the AAAI Conference on Artificial Intelligence}, volume~39, pages 25985--25993, 2025.

\bibitem{kim2024transformers}
Juno Kim and Taiji Suzuki.
\newblock Transformers provably solve parity efficiently with chain of thought.
\newblock {\em arXiv preprint arXiv:2410.08633}, 2024.

\bibitem{xu2025chain}
Silei Xu, Wenhao Xie, Lingxiao Zhao, and Pengcheng He.
\newblock Chain of draft: Thinking faster by writing less.
\newblock {\em arXiv preprint arXiv:2502.18600}, 2025.

\bibitem{qu2025tool}
Changle Qu, Sunhao Dai, Xiaochi Wei, Hengyi Cai, Shuaiqiang Wang, Dawei Yin, Jun Xu, and Ji-Rong Wen.
\newblock Tool learning with large language models: A survey.
\newblock {\em Frontiers of Computer Science}, 19(8):198343, 2025.

\bibitem{wang2023making}
Peiyi Wang, Lei Li, Liang Chen, Feifan Song, Binghuai Lin, Yunbo Cao, Tianyu Liu, and Zhifang Sui.
\newblock Making large language models better reasoners with alignment.
\newblock {\em arXiv preprint arXiv:2309.02144}, 2023.

\bibitem{lahlou2024port}
Salem Lahlou, Abdalgader Abubaker, and Hakim Hacid.
\newblock Port: Preference optimization on reasoning traces.
\newblock {\em arXiv preprint arXiv:2406.16061}, 2024.

\bibitem{xia2024beyond}
Yu~Xia, Rui Wang, Xu~Liu, Mingyan Li, Tong Yu, Xiang Chen, Julian McAuley, and Shuai Li.
\newblock Beyond chain-of-thought: A survey of chain-of-x paradigms for llms.
\newblock {\em arXiv preprint arXiv:2404.15676}, 2024.

\bibitem{baruchpreference}
Lior Baruch, Moshe Butman, Kfir Bar, and Doron Friedman.
\newblock Preference tree optimization: Enhancing goal-oriented dialogue with look-ahead simulations.
\newblock In {\em Scaling Self-Improving Foundation Models without Human Supervision}.

\bibitem{ren2025large}
Yaxuan Ren, Xufei Luo, Ye~Wang, Haodong Li, Hairong Zhang, Zeming Li, Honghao Lai, Xuanlin Li, Long Ge, Janne Estill, et~al.
\newblock Large language models in traditional chinese medicine: A scoping review.
\newblock {\em Journal of Evidence-Based Medicine}, 18(1):e12658, 2025.

\bibitem{dai2024tcmchat}
Yizheng Dai, Xin Shao, Jinlu Zhang, Yulong Chen, Qian Chen, Jie Liao, Fei Chi, Junhua Zhang, and Xiaohui Fan.
\newblock Tcmchat: A generative large language model for traditional chinese medicine.
\newblock {\em Pharmacological Research}, 210:107530, 2024.

\bibitem{wei2024biancang}
Sibo Wei, Xueping Peng, Yi-fei Wang, Jiasheng Si, Weiyu Zhang, Wenpeng Lu, Xiaoming Wu, and Yinglong Wang.
\newblock Biancang: A traditional chinese medicine large language model.
\newblock {\em arXiv preprint arXiv:2411.11027}, 2024.

\bibitem{zhang2024qibo}
Heyi Zhang, Xin Wang, Zhaopeng Meng, Zhe Chen, Pengwei Zhuang, Yongzhe Jia, Dawei Xu, and Wenbin Guo.
\newblock Qibo: A large language model for traditional chinese medicine.
\newblock {\em arXiv preprint arXiv:2403.16056}, 2024.

\bibitem{schwarcz2025ai}
Daniel Schwarcz, Sam Manning, Patrick Barry, David~R Cleveland, JJ~Prescott, and Beverly Rich.
\newblock Ai-powered lawyering: Ai reasoning models, retrieval augmented generation, and the future of legal practice.
\newblock 2025.

\bibitem{yeo2025comprehensive}
Wei~Jie Yeo, Wihan Van Der~Heever, Rui Mao, Erik Cambria, Ranjan Satapathy, and Gianmarco Mengaldo.
\newblock A comprehensive review on financial explainable ai.
\newblock {\em Artificial Intelligence Review}, 58(6):1--49, 2025.

\bibitem{tu2024findings}
L~Tu, J~Qu, S~Yavuz, S~Joty, W~Liu, C~Xiong, and Y~Zhou.
\newblock Findings of the association for computational linguistics: Eacl 2024.
\newblock In {\em Association for Computational Linguistics}, pages 1278--1294, 2024.

\bibitem{li2025aligning}
Shuyue~Stella Li, Jimin Mun, Faeze Brahman, Jonathan~S Ilgen, Yulia Tsvetkov, and Maarten Sap.
\newblock Aligning llms to ask good questions a case study in clinical reasoning.
\newblock {\em arXiv preprint arXiv:2502.14860}, 2025.

\bibitem{zhu2023ChatMed}
Wenjing~Yue Wei~Zhu and Xiaoling Wang.
\newblock Shennong-tcm: A traditional chinese medicine large language model.
\newblock \url{https://github.com/michael-wzhu/ShenNong-TCM-LLM}, 2023.

\bibitem{lin2004rouge}
Chin-Yew Lin.
\newblock Rouge: A package for automatic evaluation of summaries.
\newblock In {\em Text summarization branches out}, pages 74--81, 2004.

\bibitem{papineni2002bleu}
Kishore Papineni, Salim Roukos, Todd Ward, and Wei-Jing Zhu.
\newblock Bleu: a method for automatic evaluation of machine translation.
\newblock In {\em Proceedings of the 40th annual meeting of the Association for Computational Linguistics}, pages 311--318, 2002.

\end{thebibliography}

\end{document}